%% file: aaai21.tex
\documentclass[letterpaper]{article}
\input{command}

\title{Segatron: Segment-Aware Transformer for \\Language Modeling and Understanding}
\author{
He Bai,\textsuperscript{\rm 1}
Peng Shi,\textsuperscript{\rm 1}
Jimmy Lin,\textsuperscript{\rm 1 \rm 2}
Yuqing Xie,\textsuperscript{\rm 1}
Luchen Tan,\textsuperscript{\rm 2}
Kun Xiong,\textsuperscript{\rm 2}
Wen Gao,\textsuperscript{\rm 3}
Ming Li\textsuperscript{\rm 1 \rm 2} \\ }
\affiliations {
    \textsuperscript{\rm 1}David R. Cheriton School of Computer Science, University of Waterloo \\
    \textsuperscript{\rm 2} RSVP.ai \\
    \textsuperscript{\rm 3} School of Electronics Engineering and Computer Science, Peking University \\
    \{he.bai, peng.shi, jimmylin,  yuqing.xie, mli\}@uwaterloo.ca, \{lctan, kun\}@rsvp.ai, wgao@pku.edu.cn \\
}

\makeatletter
\patchcmd{\@startsection}{\@ifstar}{\nolinenumbers\@ifstar}{}{}
\patchcmd{\@xsect}{\ignorespaces}{\linenumbers\ignorespaces}{}{}
\makeatother

\begin{document}
\maketitle

\begin{abstract}
\input{sections/abstract}
\end{abstract}

\input{sections/introduction}

\input{sections/model}

\input{sections/experiments}

\input{sections/related_work}

\input{sections/conclusion}

\section*{Acknowledgments}
This work was partially supported by NSERC OGP0046506, the National Key R\&D Program of China 2016YFB1000902 and 2018YFB1003202.
We would like to thank Wei Zeng and his team in Peng Cheng Laboratory (PCL) for computing resources to support this project.

\bibliography{aaai21}
\clearpage

\onecolumn
\newpage
\appendix
\input{sections/appendix}

\end{document}

%% file: command.tex
\usepackage{aaai21}  
\usepackage{times}  
\usepackage{helvet} 
\usepackage{courier}  
\usepackage[hyphens]{url}  
\usepackage{graphicx} 
\usepackage{natbib}  
\usepackage{multibib}
\usepackage{caption} 
\usepackage{multirow} 
\usepackage{soul}
\usepackage{float}
\usepackage{subfigure}

\usepackage{threeparttable} 
\usepackage[utf8]{inputenc}
\usepackage[T1]{fontenc}
\usepackage[english]{babel}
\usepackage[colorlinks]{hyperref}
\usepackage{amsfonts}
\usepackage{nicefrac}
\usepackage{microtype}
\usepackage{amsmath}
\usepackage{amssymb}
\usepackage{mathtools}
\usepackage{booktabs}
\usepackage{ragged2e}
\usepackage{tikz}
\usepackage{stackengine}
\usepackage{etoolbox}
\usepackage{xspace}
\usepackage{xpatch}
\usepackage{cuted}
\usepackage{enumerate}
\usepackage{xstring}
\usepackage{setspace}
\usepackage{tabularx}
\usepackage{makecell}
\usepackage{changepage}
\usepackage{enumitem}
\usepackage{cuted}
\usepackage{titlesec}
\usepackage{cancel}
\usepackage{eqparbox}
\usepackage{bibentry}
\usepackage[hang,flushmargin]{footmisc}
\usepackage[capitalise,noabbrev,nameinlink]{cleveref}
\usepackage[useregional=numeric]{datetime2}
\usepackage[linesnumbered,ruled,noend]{algorithm2e}
\usepackage{adjustbox}

\newcites{appendix}{Appendix References}
\usepackage[switch,mathlines]{lineno}
\usepackage{lipsum}
\usepackage{etoolbox}
\usepackage{longtable}

\usetikzlibrary{positioning,arrows.meta,fit,calc,backgrounds}
\tikzset{%
node distance=2em, auto,
every node/.style={line width=0.7pt},
det/.style={draw=black, rectangle, minimum size=2.5em, inner sep=0.1ex},
lat/.style={draw=black, circle, minimum size=2.5em, inner sep=0.1ex},
obs/.style={draw=black, circle, fill=black!15, minimum size=2.5em, inner sep=0.1ex},
fac/.style={draw=black, rectangle, fill=black, minimum size=.6em, inner sep=0em},
dummy/.style={draw=none, circle, minimum size=2.5em},
plate/.style={draw=black, rounded corners, inner sep=.8em, yshift=-.7em, align=right},
box/.style={draw=black, rounded corners, inner sep=.4em, align=center},
generates/.style={->, -{Stealth[length=.6em, inset=0pt]}, line width=0.7pt},
undirected/.style={line width=0.7pt},
}

\definecolor{mydarkblue}{rgb}{0,0.08,0.45}
\hypersetup{%
colorlinks=true,
linkcolor=mydarkblue,
citecolor=mydarkblue,
filecolor=mydarkblue,
urlcolor=mydarkblue}

\graphicspath{{figures/}}

\makeatletter
\newcommand{\removeParBefore}{\ifvmode\vspace*{-\baselineskip}\setlength{\parskip}{0ex}\fi}
\newcommand{\removeParAfter}{\@ifnextchar\par\@gobble\relax}

\makeatother

\DeclareDocumentCommand{\p}{ D<>{p} D<>{} r() }{
\def\content{#3}\patchcmd{\content}{|}{\;#2\vert\;}{}{}
\ensuremath{#1 #2(\content #2)}}

\DeclareDocumentCommand{\P}{ D<>{P} D<>{\big} r() }{
\def\content{#3}\patchcmd{\content}{|}{\;#2\vert\;}{}{}
\ensuremath{\operatorname{#1}#2(\content #2)}}

\DeclareDocumentCommand{\E}{ D<>{E} D<>{\big} r[] }{
\def\content{#3}\patchcmd{\content}{|}{\;#2\vert\;}{}{}
\ensuremath{\operatorname{#1}\!#2[\content #2]}}

\DeclareDocumentCommand{\D}{ D<>{D} D<>{\big} r[] }{
\def\content{#3}\patchcmd{\content}{||}{\;#2\|\;}{}{}
\ensuremath{\operatorname{#1}\!#2[\content #2]}}

\NewDocumentCommand{\Nor}{ r() }{\P<Normal>](#1)}
\NewDocumentCommand{\Cat}{ r() }{\P<Cat>](#1)}
\NewDocumentCommand{\Bin}{ r() }{\P<Bin>](#1)}
\NewDocumentCommand{\Bet}{ r() }{\P<Beta>](#1)}
\NewDocumentCommand{\Ber}{ r() }{\P<Bernoulli>(#1)}
\NewDocumentCommand{\Dir}{ r() }{\P<Dir>(#1)}

\DeclareDocumentCommand{\KL}{ D<>{\big} r[] }{\D<KL><#1>[#2]}
\DeclareDocumentCommand{\H}{ D<>{\big} r[] }{\E<H><#1>[#2]}
\DeclareDocumentCommand{\I}{ D<>{\big} r[] }{\E<I><#1>[#2]}

%% file: sections/abstract.tex
Transformers are powerful for sequence modeling. 
Nearly all state-of-the-art language models and pre-trained language models are based on the Transformer architecture. 
However, it distinguishes sequential tokens only with the token position index.
We hypothesize that better contextual representations can be generated from the Transformer with richer positional information.
To verify this, we propose a segment-aware Transformer~(Segatron), by replacing the original token position encoding with a combined position encoding of paragraph, sentence, and token. 
We first introduce the segment-aware mechanism to Transformer-XL, which is a popular Transformer-based language model with memory extension and relative position encoding. 
We find that our method can further improve the Transformer-XL base model and large model, achieving 17.1 perplexity on the WikiText-103 dataset.
We further investigate the pre-training masked language modeling task with Segatron. 
Experimental results show that BERT pre-trained with Segatron~(SegaBERT) can outperform BERT with vanilla Transformer on various NLP tasks, and outperforms RoBERTa on zero-shot sentence representation learning. Our code is available on GitHub.\footnote{\url{https://github.com/rsvp-ai/segatron_aaai}}

%% file: sections/introduction.tex
\section{Introduction}

Language modeling~(LM) is a traditional sequence modeling task which requires learning long-distance dependencies for next token prediction based on the previous context.
Recently, large neural LMs trained on a massive amount of text data have shown great potential for representation learning and transfer learning, and also achieved state-of-the-art results in various natural language processing tasks. 

To the best of our knowledge, state-of-the-art language models~\cite{DBLP:conf/acl/DaiYYCLS19, DBLP:conf/iclr/BaevskiA19, DBLP:conf/iclr/RaePJHL20} and pre-trained language models~\cite{gpt,DBLP:journals/corr/bert,DBLP:conf/nips/XLNet,DBLP:conf/iclr/albert} all use a multi-layer Transformer~\cite{DBLP:conf/nips/VaswaniSPUJGKP17}. 
The Transformer network was initially used in the seq2seq architecture for machine translation, whose input is usually a sentence. 
Hence, it is intuitive to distinguish each token with its position index in the input sequence. 
However, the input length can grow to 1024 or more tokens and come from different sentences and paragraphs for language modeling. 
Although vanilla position encoding can help the transformer be aware of the token position by assigning a unique index to each token, the token index in a sentence, sentence index in a paragraph, and paragraph index in a document are all implicit.
Such segmentation information is essential for language modeling, as tokens in different segments of context hold different significance for next token prediction.
If the Transformer model can be aware of the segment position of each context token, we hypothesize that better context representations will be encoded.
This statement is not made lightly, as for 3000 years, many languages including ancient Latin, Greek, English, French, and Chinese did not have punctuations or paragraphs. 
The introduction of sentence and paragraph separators was fundamental, so is indexing them to train Transformers. 
Although punctuations and paragraph breakers can provide boundary information to some extent, the boundary is not as straightforward as segment position, especially for the dot-product self-attention based Transformer.

Hence, we propose a novel segment-aware Transformer~(Segatron), which encodes paragraph index in a document, sentence index in a paragraph, and token index in a sentence all together for the input sequence. 
We first verify the proposed method with relative position encoding on the language modeling task. 
By applying the segment-aware mechanism to Transformer-XL~\cite{DBLP:conf/acl/DaiYYCLS19}, our base model trained with the WikiText-103 dataset~\cite{DBLP:conf/iclr/MerityX0S17} outperforms Transformer-XL base by 1.5 points in terms of perplexity. 
Our large model achieves a perplexity of 17.1, the same score as Compressive Transformer~\cite{DBLP:conf/iclr/RaePJHL20}, which is a more complicated model with longer input context and additional training objectives.
We also pre-train masked language models with Transformer~(BERT-base) and Segatron~(SegaBERT-base) with English Wikipedia for 500K training steps. 
According to experimental results, SegaBERT can outperform BERT on both general language understanding~(GLUE) and machine reading comprehension~(SQUAD and RACE) tasks.
We further pre-trained a large model SegaBERT-large with the same data used in BERT. Experimental results show that SegaBERT-large can not only outperform BERT-large on all the above tasks, but also outperforms RoBERTa-large on zero-shot Semantic Textual Similarity tasks.
These results demonstrate the value of segment encodings in Transformers.

%% file: sections/model.tex
\section{Model}
In this section, we show how to apply our proposed segment-aware Transformer to language modeling. More specifically, we first introduce our Segatron-XL (Segment-aware Transformer-XL) with non-learnable relative position encoding for autoregressive language modeling. 
Then we introduce our pre-trained Segatron~(SegaBERT) with learnable absolute position encoding for masked language modeling~(MLM).

\subsection{Segatron-XL}
We first introduce our method in the context of autoregressive language modeling, by replacing the vanilla Transformer index in  Transformer-XL~\citep{DBLP:conf/acl/DaiYYCLS19} with Segatron.
Transformer-XL is a memory augmented Transformer with relative position encoding: 
\begin{equation}\label{equ:attn_tranxl}
  \begin{array}{ll}
    \mathbf{A}_{i,j}^{rel}&= \mathbf{E}_{x_i}^T\mathbf{W}_q^T\mathbf{W}_{k,E}\mathbf{E}_{x_j} + \mathbf{E}_{x_i}^T\mathbf{W}_q^T\mathbf{W}_{k,R}\mathbf{R}_{i-j} \\
    & +  u^T\mathbf{W}_{k,E}\mathbf{E}_{x_j} +  v^T\mathbf{W}_{k,R}\mathbf{R}_{i-j}
  \end{array}
\end{equation} 
where $\mathbf{A}_{i,j}^{rel}$ is the self-attention score between query $i$ and key $j$. 
$\mathbf{E}_{x_i}$ and $\mathbf{E}_{x_j}$ are the input representations of query $i$ and key $j$, respectively. 
$\mathbf{R}_{i-j}$ is the relative position embedding. 
$\mathbf{W}_{k,E}$ and $\mathbf{W}_{k,R}$ are transformation matrices for input representation and position embedding, respectively. 
$\mathbf{u}$ and  $\mathbf{v}$ are learnable variables.
The position embeddings are non-learnable and defined as:
\begin{equation}\label{equ:pos_tranxl}
  \mathbf{R}_{i-j,k} = \left\{
  \begin{array}{ll}
    sin(\frac{i-j}{10000^{2k/dim}})  & k<\frac{1}{2}dim \\
    cos(\frac{i-j}{10000^{2k/dim}})  &  k \geq \frac{1}{2}dim
  \end{array}
\right.
\end{equation} 
where $dim$ is the dimension size of $\mathbf{\mathbf{R}_{i-j}}$, and $k$ is the dimension index. 

Our proposed method introduces paragraph and sentence segmentation to the relative position encoding. The new position embeddings $\mathbf{R}_{\mathbf{I},\mathbf{J}}$ are defined as:
\begin{equation}\label{equ:pos_segatranxl}
  \mathbf{R}_{\mathbf{I},\mathbf{J},k} = \left\{
  \begin{array}{ll}
    \mathbf{R^t}_{t_i-t_j,k}  &k<\frac{1}{3}dim \\
    \mathbf{R^s}_{s_i-s_j,k-\frac{1}{3}dim}  &\frac{2}{3}dim>k \geq \frac{1}{3}dim \\
    \mathbf{R^p}_{p_i-p_j,k-\frac{2}{3}dim}  &k \geq \frac{2}{3}dim 
  \end{array}
  \right.
\end{equation} 
where $\mathbf{I}=\{t_i,s_i,p_i\}$, $\mathbf{J}=\{t_j,s_j,p_j\}$. 
$t$, $s$, and $p$ are token position index, sentence position index, and paragraph position index, respectively. 
$\mathbf{R^t}$, $\mathbf{R^s}$, and $\mathbf{R^p}$ are the relative position embeddings of token, sentence, and paragraph. 
These embeddings are defined in Eq.~\ref{equ:pos_tranxl} and the dimensions of each are equal to 1/3 of $\mathbf{R}_{\mathbf{I},\mathbf{J}}$.
The input representation of our model is shown in Figure~\ref{fig:architecture-segatronxl}.

To equip the recurrence memory mechanism of Transformer-XL with the segment-aware relative position encoding, the paragraph position, the sentence position, and the token position indexes of the previous segment should also be cached together with the hidden states. Then, the relative position can be calculated by subtracting the cached position indexes from the current position indexes.

\input{figs/arichitecture.tex}
\subsection{Pre-trained Segatron}
\label{section: model/segabert}
We will introduce how to pre-train a language model with our proposed Segatron in this section.

First, pre-training a masked language model in the setting of BERT is a practical choice, as BERT is a popular baseline model and requires less computational resources compared with more recent large models. 
For example, BERT-large only needs about 10\% of the resources of RoBERTa-large~\cite{DBLP:journals/corr/abs-1907-11692}.
Hence, in this paper, we first pre-train two base size models: $\text{SegaBERT-base}^{-}$ and $\text{BERT-base}^{-}$ with only English Wikipedia data for 500K training steps, to compare BERT pre-trained with Transformer and Segatron fairly. 
We then pre-train a large size model SegaBERT-large with Wikibooks dataset and 1M training steps, same as BERT-large. 

\smallskip\noindent\textbf{Input Representation.} 
Input $\mathbf{X}$ of SegaBERT is a sequence of tokens, which can be one or more sentences or paragraphs.
The representation $x_t$ for token $t$ is computed by summing the corresponding token embedding $\mathbf{E}_t$, token index embedding $\mathbf{P}^t_t$, sentence index embedding $\mathbf{P}^s_t$, and paragraph index embedding $\mathbf{P}^p_t$, as shown in Figure~\ref{fig:architecture-segabert}.
Two special tokens \texttt{[CLS]} and \texttt{[SEP]} are added to the text sequence before the first token and after the last token, and their paragraph/sentence indexes are the same as their adjacent tokens.
Following BERT, the text is tokenized into subwords with WordPiece and the maximum sequence length is 512. 

\smallskip\noindent\textbf{Training Objective.}
Following BERT, we use the masked LM as our training objective.
However, next sentence prediction (NSP) is not used in our model, as our input contains more than two sentences.

\input{tables/segatronxl-wiki103}
\smallskip\noindent\textbf{Data preparation.}
For the pre-training corpus we use English Wikipedia and Bookcorpus~\cite{DBLP:conf/iccv/bookcorpus}.
For each document, we firstly split each into $N_p$ paragraphs, and all the sub-tokens in the $i$-th paragraph are assigned the same Paragraph Index Embedding $\mathbf{P}^p_{i}$.
The paragraph index starts from 0 for each document.
Similarly, each paragraph is further segmented into $N_s$ sentences with NLTK~\cite{DBLP:books/daglib/nltk}, and all the sub-tokens in the $i$-th sentence are assigned the same Sentence Index Embedding $\mathbf{P}^s_{i}$.
The sentence index starts from 0 for each paragraph.
Within each sentence, all the sub-tokens are indexed from 0; 
the $i$-th sub-token will have its Token Index Embedding $\mathbf{P}^t_i$.

When building a training example, we randomly~(length weighted) sample a document from the corpus and randomly select a sentence in that document as the start sentence. 
Then, the following sentences are added to that example until the example meets the maximum length limitation~(512) or runs out of the selected document.
If any position index in that example exceeds the maximum index, all such position indexes will be subtracted by one until they meet the maximum requirements.
The maximum position index of paragraph, sentence, and token are 50, 100, and 256, respectively.

\smallskip\noindent\textbf{Training Setup.}
\citet{DBLP:journals/corr/abs-1907-11692} have shown that BERT pre-trained with document input (more than two sentences) without NSP performs better than the original BERT on some tasks.
Hence, we not only pre-train a SegaBERT-large, but also pre-train two base models with the same setting for fair comparison.
Similar to BERT, the base model is 12 layers, 768 hidden size, and 12 self-attention heads. The large model is 24 layers, 1024 hidden size, and 24 self-attention heads. 
For optimization, we use Adam with learning rate 1e-4, $\beta_1$=0.9, $\beta_2$=0.999, with learning rate warm-up over the first 1\% of the total steps and with linear decay of the learning rate.

%% file: figs/arichitecture.tex
\begin{figure}[t]
  \centering
  \subfiguretopcaptrue
  \subfigure[Concating Relative Position Embedding]{\includegraphics[width=0.4\textwidth]{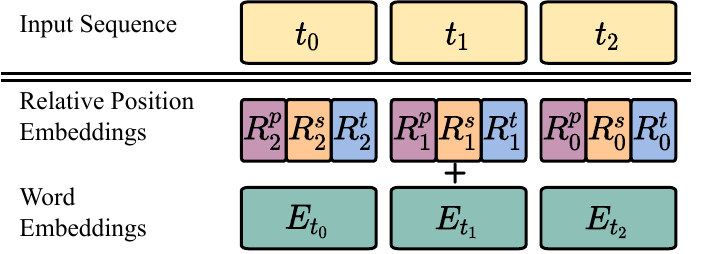}\label{fig:architecture-segatronxl}}
  \subfigure[Summing Absolute Position Embedding]{\includegraphics[height=0.22\textwidth ]{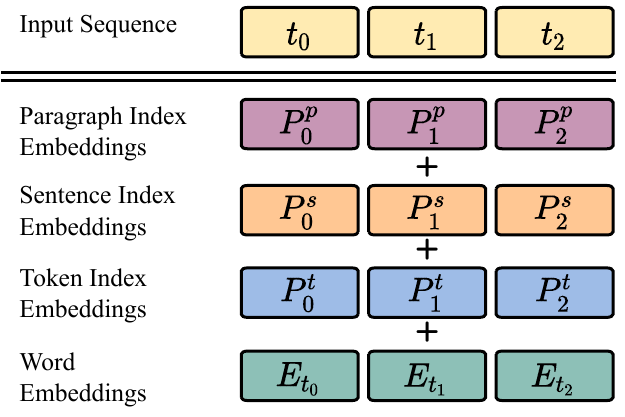}\label{fig:architecture-segabert}}
  
  \caption{Input representation of Segatron-XL and SegaBERT.}
  \label{fig:architecture}
\end{figure}

%% file: tables/segatronxl-wiki103.tex
\begin{table*}[t]\centering
  \small
  \begin{tabular}{lcc}\toprule
  Model &\#Param. &PPL \\\midrule
  LSTM+Neural cache~\cite{DBLP:conf/iclr/GraveJU17} &- &40.8 \\
  Hebbian+Cache~\cite{DBLP:conf/icml/RaeDDL18} &- &29.9 \\
  Transformer-XL base, M=150~\cite{DBLP:conf/acl/DaiYYCLS19} &151M &24.0 \\
  Transformer-XL base, M=150~(ours) &151M &24.4 \\
  Segatron-XL base, M=150 &151M &\textbf{22.5} \\
  \hline
  Adaptive Input~\cite{DBLP:conf/iclr/BaevskiA19} & 247M &18.7 \\
  Transformer-XL large, M=384~\cite{DBLP:conf/acl/DaiYYCLS19} &257M &18.3 \\
  Compressive Transformer, M=1024~\cite{DBLP:conf/iclr/RaePJHL20} &257M&17.1 \\
  Segatron-XL large, M=384 &257M &\textbf{17.1} \\
  \bottomrule
  \end{tabular}
  \caption{Comparison with Transformer-XL and competitive baseline results on WikiText-103.}\label{tab: wiki103}
\end{table*}

%% file: sections/experiments.tex
\section{Experiments}
In this section, we first conduct autoregressive language modeling experiments with our proposed Segatron and also conduct an ablation study with this task.
Then, we show the results of pre-trained SegaBERT on general language understanding tasks, semantic textual similarity tasks, and machine reading comprehension tasks.

\subsection{Autoregressive Language Modeling}

\subsubsection{Dataset}
WikiText-103 is a large word-level dataset with long-distance dependencies for language modeling. 
This dataset preserves both punctuations and paragraph line breakers, which are essential for our segmentation pre-processing.
There are 103M tokens, 28K articles for training. 
The average length is 3.6K tokens per article. 

\subsubsection{Model Configuration}
Following Transformer-XL, we train a base size model and a large size model.
The base model is a 16 layer Transformer with a hidden size of 410 and 10 self-attention heads. This model is trained for 200K steps with a batch size of 64. 
The large model is an 18 layer Transformer with a hidden size of 1024 and 16 attention heads. 
This model is trained with 350K steps with a batch size of 128. 
The sequence length and memory length during training and testing all equal 150 for the base model and 384 for the large model. 
The main differences between our implementation and Transformer-XL are:\ we use mixed-precision mode; our input/memory lengths between training and testing are the same; the large model training steps of Transformer-XL are 4M according to their implementation.

\subsubsection{Main Results}
Our results are shown in Table~\ref{tab: wiki103}. 
As we can see from this table, the improvement with the segment-aware mechanism is quite impressive:\ the perplexity decreases 1.5 points for the Transformer-XL base and decreases 1.2 for Transformer-XL large. 
We also observe that our large model achieves 18.3 PPL with only 172K training steps.
We finally obtain a perplexity of 17.1 with our large model -- comparable to prior state-of-the-art results of Compressive Transformer~\cite{DBLP:conf/iclr/RaePJHL20}, which is based on Transformer-XL but trained with longer input length and memory length~(512) and a more complicated memory cache mechanism.

\input{figs/segatronxl_ppl.tex}
\input{figs/segatronxl_length.tex}
\input{tables/segatronxl-ablation}

It is worth noting that we do not list methods with additional training data or dynamic evaluation~\cite{DBLP:conf/icml/KrauseK0R18} which continues training the model on the test set. 
We also note that there is a contemporaneous work RoutingTransformer~\cite{DBLP:journals/corr/abs-2003-05997}, which modifies the self-attention to local and sparse attention with a clustering method. 
However, their implementations are not available.
We believe our method is orthogonal to their work and can be introduced to their model.

\subsubsection{Analysis}
We plot the valid perplexity of Segatron-XL base and Transformer-XL base during training in Figure~\ref{fig: ppl_wiki103}. 
From this figure, we can see that the segment-aware model outperforms the base model all the time, and the gap between them becomes larger as training progresses. 
Segatron-XL at 10K steps approximately matches the performance of Transformer-XL at 20K steps.
We then test the effectiveness of Segatron over different input lengths (25, 50, 100, and 150 input tokens) by comparing Transformer-XL and Segatron-XL base models.
As we can see from Figure~\ref{fig: segatronxl_length}, the improvements are consistent and significant. 
There is no evidence showing our method prefers shorter or longer input.

\input{figs/segabert_ppl.tex}

\subsubsection{Ablation Study}
We finally conduct an ablation study with Segatron-XL base, to investigate the contributions of the sentence position encoding and the paragraph position encoding, respectively. 
Experimental results are shown in Table~\ref{tab:ablation_ppl}.
From this table, we find that the PPL of Transformer-XL decreases from 24.35 to 24.07/22.51 after adding paragraph/sentence position encoding, and further decreases to 22.47 by encoding paragraph and sentence positions simultaneously.
The results show that both the paragraph position and sentence position can help the Transformer to model language. 
Sentence position encoding contributes more than paragraph position encoding in our experiments. 

\input{tables/glue_fair_com.tex}
\input{tables/glue_test_com.tex}
\input{tables/sts-zero_shot.tex}

\subsection{Pre-trained Masked Language Model}
We first plot the valid losses of $\text{BERT-base}^{-}$ and $\text{SegaBERT-base}^{-}$ during pre-training in Figure~\ref{fig: ppl_pretrain}.
The overall trends between Figure~\ref{fig: ppl_wiki103} and Figure~\ref{fig: ppl_pretrain} are similar, which demonstrates that our proposed segment-aware method works on both auto-regressive language modeling and masked language modeling.
We will detail our experiments with our pre-trained models in the following sections.

\subsubsection{General Language Understanding}
The General Language Understanding Evaluation~(GLUE) benchmark~\cite{DBLP:conf/iclr/glue} is a collection of resources for evaluating natural language understanding systems. 
Following \citet{DBLP:journals/corr/bert}, we evaluate our model over these tasks: linguistic acceptability CoLA~\cite{cola}, sentiment SST-2~\cite{sst2}, paraphrase MRPC~\cite{mrpc}, textual similarity STS-B~\cite{sts-b}, question paraphrase QQP, textual entailment RTE~\cite{rte} and MNLI~\cite{mnli}, and question entailment QNLI~\cite{DBLP:conf/iclr/glue}. 
We fine-tune every single task only on its in-domain data without two-stage transfer learning.

On the GLUE benchmark, we conduct the fine-tuning experiments in the following manner: For single-sentence classification tasks, such as sentiment classification~(SST-2), the sentence will be assigned Paragraph Index 0 and Sentence Index 0.
For sentence pair classification tasks, such as question-answer entailment~(QNLI), the first sentence will be assigned Paragraph Index 0 and Sentence Index 0 and the second sentence will be assigned Paragraph Index 1 and Sentence Index 0. 

We conduct grid search with the GLUE dev set for small data tasks:\ CoLA, MRPC, RTE, SST-2, and STS-B. 
Our grid search space is as follows: 

\begin{itemize}
    \item Batch size: 16, 24, 32;  
    \item Learning rate: 2e-5, 3e-5, 5e-5; 
    \item Number of epochs: 3-10.
\end{itemize}

For QQP, MNLI, and QNLI, we use the default hyper-parameters:\ 3e-5 learning rate, 256 batch size, and 3 epochs. 
The other hyper-parameters are the same as in the HuggingFace Transformers library.\footnote{\url{https://github.com/huggingface/transformers}} 

We compare BERT and SegaBERT in a fair setting to decouple the effects of document-level inputs and the removal of NSP.
In Table~\ref{tab: glue_fair}, two base models are pre-trained by us and the only difference is the position encoding.
We can see that our $\text{SegaBERT-base}^{-}$ outperforms $\text{BERT-base}^{-}$ on most tasks. 
We also notice that $\text{SegaBERT-base}^{-}$ is lower than $\text{BERT-base}^{-}$ by over 2.5 points on CoLA. 
However, this gap decreases to 0.1 on the test set, which is shown in Table~\ref{tab: glue_test}.
This is because the size of CoLA is quite small and not as robust as other datasets.
Improvements can also be observed easily when comparing SegaBERT-large with the best score of 3 BERT-large models. 

These results demonstrate SegaBERT's effectiveness in general natural language understanding. 
The improvements on these sentence and sentence pair classification tasks show that our segment-aware pre-trained model is better than vanilla Transformer on sentence-level tasks.

\subsubsection{Sentence Representation Learning}
Since our SegaBERT has shown great potential on sentence-level tasks, in this section, we further investigate whether SegaBERT can generate better sentence representations.
Following Sentence-BERT~\citep{DBLP:conf/emnlp/sbert}, we fine-tune SegaBERT in a siamese structure on the combination of SNLI~\cite{DBLP:conf/emnlp/BowmanAPM15} and MNLI datasets. The fine-tuned model is named S-SegaBERT.
We then evaluate the zero-shot performance of S-SegaBERT and other baselines on Semantic Textual Similarity (STS) tasks using the Spearman's rank correlation between the cosine similarity of the sentence embeddings and the gold labels. 

In Table~\ref{tab: sts-zeroshot}, the results of S-BERT-large and S-RoBERTa-large are from \citet{DBLP:conf/emnlp/sbert}. 
The results of S-BERT-large* are re-implemented by us, which is similar to Sentence-BERT's results.
We can see that our SegaBERT achieves the highest average scores on STS tasks, even outperforms RoBERTa, which uses much more training data, larger batch size, and dynamic masking.
These results conform with our improvements on GLUE benchmarks, which indicate that a language model pre-trained with Segatron can learn better sentence representations~(single sentence encoding) than the original Transformer.

\input{figs/attention.tex}

\subsubsection{Reading Comprehension}
We finally test our pre-trained model on machine reading comprehension tasks.
For these tasks, the question is assigned Paragraph Index 0 and Sentence Index 0.
For a context with $n$ paragraphs, Paragraph Index 1 to $n+1$ are assigned to them accordingly.
Within each paragraph, the sentences are indexed from 0.

\input{tables/squad.tex}

We first fine-tune our SegaBERT model with SQUAD v1.1~\cite{DBLP:conf/emnlp/squad11} for 4 epochs with 128 batch size and 3e-5 learning rate. The fine-tuning setting of SQUAD v2.0~\cite{DBLP:conf/acl/squad2} is the same as SQUAD v1.1. Results are shown in Table~\ref{tab: squad}. 
As we can see from Table~\ref{tab: squad}, our pre-trained $\text{SegaBERT-base}^{-}$ outperforms our pre-trained $\text{BERT-base}^{-}$ on both dataset:\ 1.3 EM and 0.8 F1 improvements on SQUAD v1.1; 0.9 EM and 1.0 F1 improvements on SQUAD v2. 
It should be noticed that our pre-trained $\text{BERT-base}^{-}$ outperforms the original BERT-base model, although ours is pre-trained with fewer data and steps. 
This confirms~\citet{DBLP:journals/corr/abs-1907-11692}'s finding that BERT pre-trained with document-level input can contribute to performance improvements on SQUAD.
For large models, as we cannot afford to train a new BERT-large model in the same setting as $\text{BERT-base}^{-}$, we compare our model with BERT-large wwm~(with whole word masking), which is a stronger baseline model.
We can see that SegaBERT large is slightly lower than BERT-large wwm on SQUAD v1.1 but outperforms it on SQUAD v2 over 1.2 EM and 1.8 F1.

\input{tables/race.tex}

We further test our models with RACE~\citep{DBLP:conf/emnlp/LaiXLYH17}, which is a large-scale reading comprehension dataset with more than 28,000 passages. 
RACE has significantly longer contexts than SQUAD.
Our results are shown in Table~\ref{tab: race}.
The overall trend is similar to SQUAD.


\subsubsection{Visualization}

We further visualize the self-attention scores of $\text{BERT-base}^{-}$ and $\text{SegaBERT-base}^{-}$ in different layers. 
Figure~\ref{fig:heatmap} shows the average attention scores across different attention heads. 
By comparing Figure~\ref{fig: sega 1} with Figure~\ref{fig: bert 1}, we find that SegaBERT can capture context according to the segmentation, for example, tokens tend to attend more to tokens in its paragraph than tokens in the other paragraphs. 
A similar trend can be observed at the sentence level but is more prominent in the shallow layers
On the other hand, the BERT model seems to pay more attention to its neighbors:\ the attention weights of the elements around the main diagonal are larger than other positions in Figure~\ref{fig: bert 1}, and a band-like contour around the main diagonal can be observed in this figure.

From Figure~\ref{fig: sega 12} and Figure~\ref{fig: bert 12}, we can see the attention structure in the final layer is different from the shallow layers, and SegaBERT pays more attention to its context than BERT.
We also notice that a fractal-like structure can be observed in the first 10 layers of SegaBERT, while the last two layers of SegaBERT have a striped structure.

These attention behaviors show that:\ in the shallow layers, our model is segment-aware while BERT is neighborhood-aware; 
in the top layers, both of these two models focus on some tokens across the article rather than local neighbors, but our model can capture more contextual tokens.

%% file: figs/segatronxl_ppl.tex
\begin{figure}[t]
  \centering
  \includegraphics[width=0.35\textwidth,height=0.3\textwidth ]{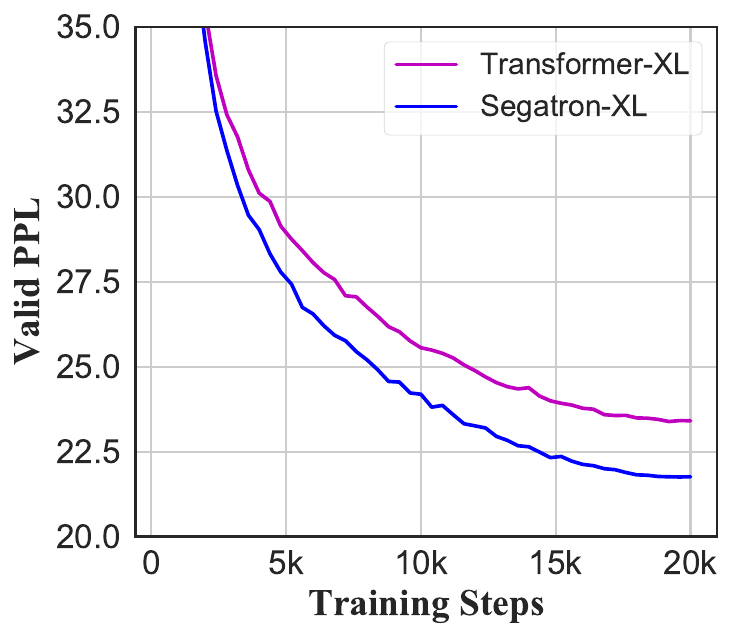}
  \caption{Valid perplexities during the training processes of language modeling.}
  \label{fig: ppl_wiki103}
\end{figure}

%% file: figs/segatronxl_length.tex
\begin{figure}[ht!]
  \centering
  \includegraphics[width=0.35\textwidth]{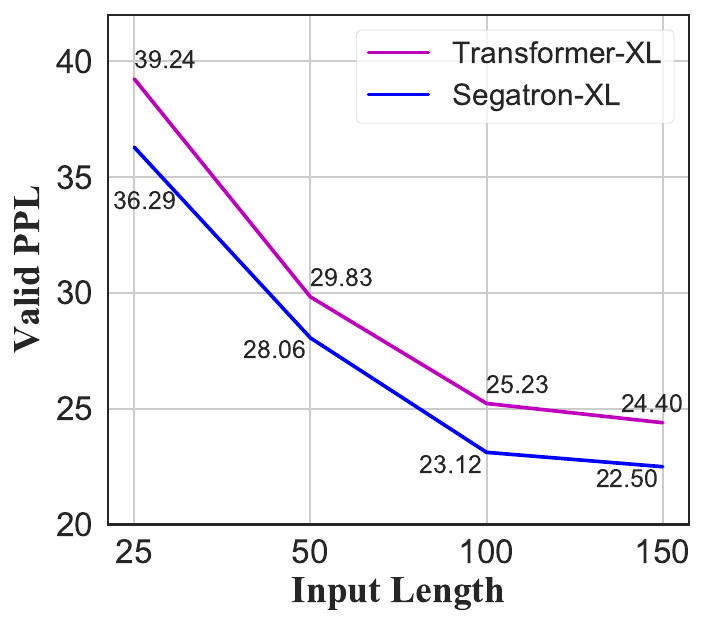}
  \caption{Test perplexities of Segatron-XL and Transformer-XL with different input lengths.}
  \label{fig: segatronxl_length}
\end{figure}

%% file: tables/segatronxl-ablation.tex
\begin{table}[t]\centering
  \begin{tabular}{lc}\toprule
  Model &PPL \\\midrule
  Transformer-XL base &24.35 \\
  ~+ paragraph position encoding &24.07 \\
  ~+ sentence position encoding &22.51 \\
  Segatron-XL base &22.47 \\
  \bottomrule
  \end{tabular}
  \caption{Ablation over the position encodings using Transformer-XL base architecture.}\label{tab:ablation_ppl}
  \end{table}

%% file: figs/segabert_ppl.tex
\begin{figure}[t]
  \centering
  \includegraphics[width=0.38\textwidth,height=0.3\textwidth ]{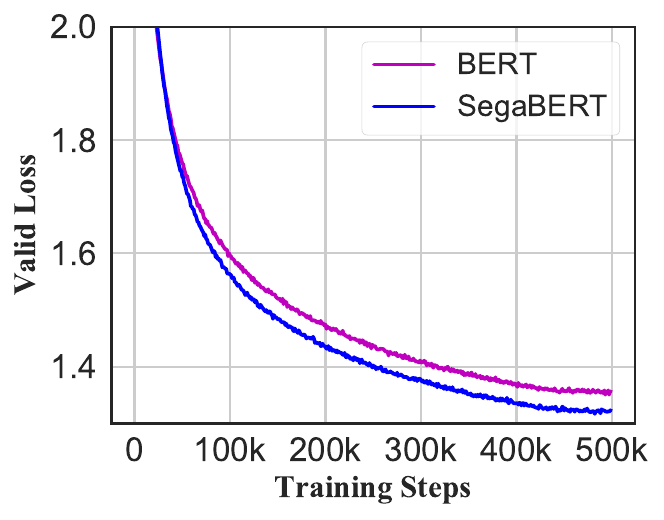}
  \caption{Valid losses during the pre-training.}
  \label{fig: ppl_pretrain}
\end{figure}

%% file: tables/glue_fair_com.tex
\begin{table*}[!htp]\centering
  \small
  \begin{tabular}{lrrrrrrrrrr}\toprule
    Model &MNLI &QNLI &QQP &RTE &SST-2 &MRPC &CoLA &STS-B &AVG \\\midrule
    $\text{BERT-base}^{-}$ &83.2 &90.4 &86.5 &68.3 &91.3 &\textbf{92.6} &55.0 &88.9 &82.0 \\
    $\text{SegaBERT-base}^{-}$ &83.8 &91.5 &87.0 &71.8 &92.1 &92.4 &54.7 &89.0 &82.8 \\
    BERT-large~(best of 3) &87.3 &93.0 &\textbf{91.4} &74.0 &94.0 &88.7 &63.7 &90.2 &85.3 \\
    SegaBERT-large &\textbf{87.6} &\textbf{93.6} &89.1 &\textbf{78.3} &\textbf{94.7} &92.3 &\textbf{65.3} &\textbf{90.3} &\textbf{86.4} \\
  \bottomrule
  \end{tabular}
  \caption{Fair comparison on GLUE dev. The two base models are pre-trained in the same setting. For large models comparison, we choose the best of 3 BERT-large models:\ the original BERT, whole word masking BERT, and BERT without NSP task. Results of BERT-large~(best of 3) are from~\citet{DBLP:conf/nips/XLNet}.}\label{tab: glue_fair}
  \end{table*}

%% file: tables/glue_test_com.tex
\begin{table*}[ht]\centering
  \small
  \begin{tabular}{lrrrrrrrrrr}\toprule
    Model &MNLI &QNLI &QQP &RTE &SST-2 &MRPC &CoLA &STS-B &AVG \\\midrule
    $\text{BERT-base}^{-}$ &82.9 &90.1 &70.8 &65.4 &91.2 &88.9 &43.5 &83.9 &77.1 \\
    $\text{SegaBERT-base}^{-}$ &83.5 &90.8 &71.4 &68.1 &91.5 &89.3 &50.7 &84.6 &78.7 \\
    BERT-large &86.7 &92.7 &72.1 &70.1 &\textbf{94.9} &89.3 &60.5 &86.5 &81.6 \\
    SegaBERT-large &\textbf{87.9} &\textbf{94.0} &\textbf{72.5} &\textbf{71.6} &94.8 &\textbf{89.7} &\textbf{62.6} &\textbf{88.6} &\textbf{82.7} \\
  \bottomrule
  \end{tabular}
  \caption{Results on GLUE test set. Results of BERT-large are from~\citet{DBLP:journals/corr/bert}.}\label{tab: glue_test}
  \end{table*}

%% file: tables/sts-zero_shot.tex
\begin{table*}[ht!]\centering
  \small
  \begin{tabular}{lrrrrrrrrr}\toprule
   Model&STS-12 &STS-13 &STS-14 &STS-15 &STS-16 &STS-B &SICK-R &AVG \\\midrule
  S-BERT-large &72.27 &78.46 &74.90 &80.99 &76.25 &79.23 &73.75 &76.55 \\
  S-BERT-large* &72.39 &78.06 &\textbf{75.26} &81.79 &76.35 &78.64 &73.85 &76.62 \\
  S-RoBERTa-large &\textbf{74.53} &77.00 &73.18 &81.85 &76.82 &79.10 &\textbf{74.29} &76.68 \\
  \hline
  S-SegaBERT-large &74.49 &\textbf{78.64} &74.88 &\textbf{83.28} &\textbf{77.10} &\textbf{79.42} &73.77 &\textbf{77.37} \\  \bottomrule
  \end{tabular}
  \caption{Zero-shot spearman's rank correlation $\rho\times 100$ between the negative distance of sentence embeddings and the gold labels. STS-B: STS benchmark, SICK-R: SICK relatedness dataset. Results of BERT-large and RoBERTa-large are from~\citet{DBLP:conf/emnlp/sbert}.}\label{tab: sts-zeroshot}
  \scriptsize
  \end{table*}

%% file: figs/attention.tex
\begin{figure*}[t]
  \centering

  \subfigure[BERT-Layer 1]{\includegraphics[width=0.27\textwidth]{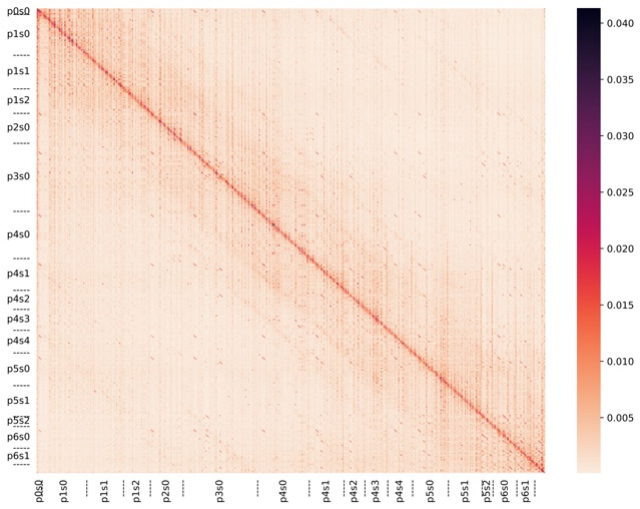}\label{fig: bert 1}}
  \subfigure[BERT-Layer 6]{\includegraphics[width=0.27\textwidth]{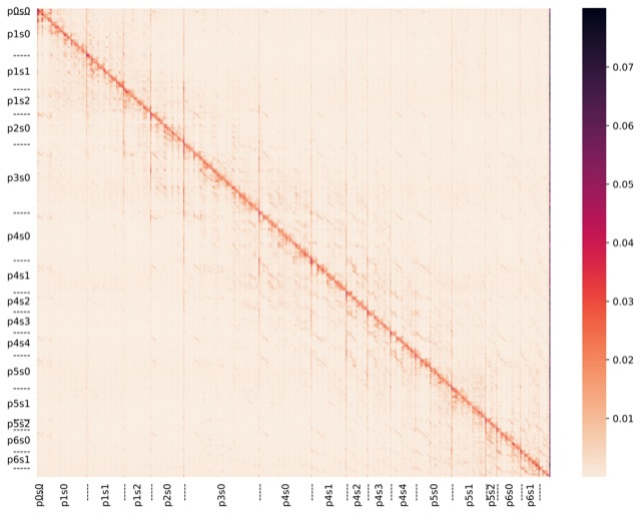}\label{fig: bert 6}}
  \subfigure[BERT-Layer 12]{\includegraphics[width=0.27\textwidth]{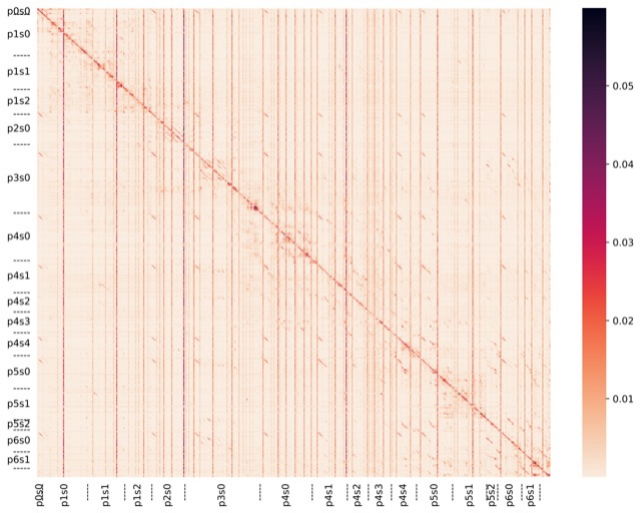}\label{fig: bert 12}}
  \subfigure[SegaBERT-Layer 1]{\includegraphics[width=0.27\textwidth]{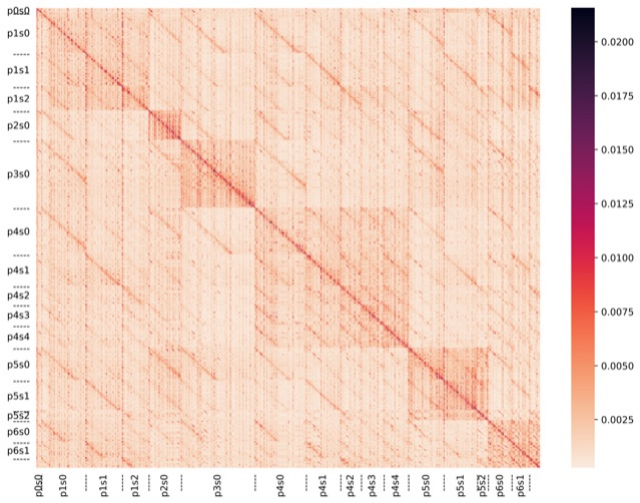}\label{fig: sega 1}}
  \subfigure[SegaBERT-Layer 6]{\includegraphics[width=0.27\textwidth]{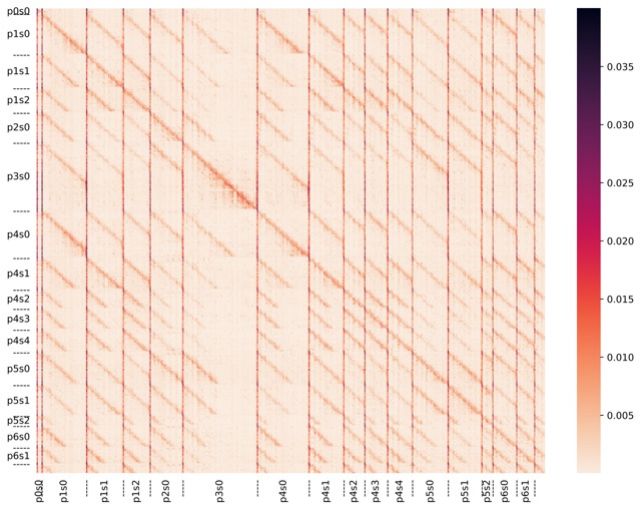}\label{fig: sega 6}}
  \subfigure[SegaBERT-Layer 12]{\includegraphics[width=0.27\textwidth]{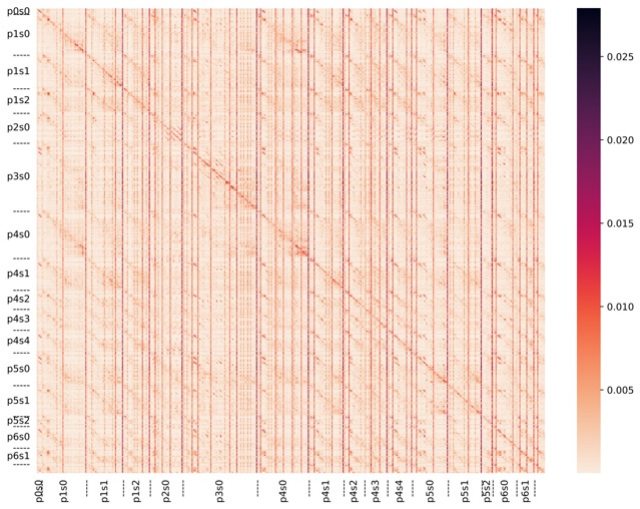}\label{fig: sega 12}}

  \caption{Self-attention heat maps of the first, the sixth, and the last layer of SegaBERT and BERT when encoding the first 512 tokens of a Wikipedia article.}
  \label{fig:heatmap}
\end{figure*}

%% file: tables/squad.tex
\begin{table}[t]\centering
  \small
  \begin{tabular}{lrrrrr}\toprule
  \multirow{2}{*}{System} &\multicolumn{2}{c}{SQUAD1.1} &\multicolumn{2}{c}{SQUAD2.0} \\\cmidrule{2-5}
  Model&EM &F1 &EM &F1 \\\midrule
  BERT-base &80.8 &88.5 &72.3 &75.6 \\
  $\text{BERT-base}^{-}$ &81.9 &89.4 &75.4 &78.2 \\
  $\text{SegaBERT-base}^{-}$ &83.2 &90.2 &76.3 &79.2 \\
  BERT-large &84.1 &90.9 &78.7 &81.9 \\
  BERT-large wwm &86.7 &92.8 &80.6 &83.4 \\
  SegaBERT-large &86.0 &92.6 &81.8 &85.2 \\
  \bottomrule
  \end{tabular}
  \caption{Evaluation results on SQUAD v1.1 and v2. Results of BERT-base and BERT-large are from~\citet{DBLP:journals/corr/bert}. Results of BERT-large wwm on SQUAD v1.1 are from BERT's github repository. There are no official results of BERT-large wwm on SQUAD v2 and here we report our fine-tuning results.}

  \label{tab: squad}
  \end{table}

%% file: tables/race.tex
\begin{table}[t]\centering
  \small
  \begin{tabular}{lccc}\toprule
  Model&Acc-Dev &Acc-Test \\\midrule
  BERT-large &72.7 &72.0 \\
  SegaBERT-large &74.5 &73.8 \\
  \bottomrule
  \end{tabular}
  \caption{Accuracy on dev and test sets of RACE. Results of BERT-large are from~\citet{DBLP:conf/acl-mrqa/PanSYCJCY19}.}
  \label{tab: race}
  \end{table}

%% file: sections/related_work.tex
\section{Related Work}
Language modeling is a traditional natural language processing task which requires capturing long-distance dependencies for predicting the next token based on the context. 

Most of the recent advances in language modeling are based on the Transformer~\cite{DBLP:conf/nips/VaswaniSPUJGKP17} decoder architecture.
\citet{DBLP:conf/aaai/Al-RfouCCGJ19} demonstrated that self-attention can perform very well on character-level language modeling. 
\citet{DBLP:conf/iclr/BaevskiA19} proposed adaptive word input representations for the Transformer to assign more capacity to frequent words and reduce the capacity for less frequent words. 
\citet{DBLP:conf/acl/DaiYYCLS19} proposed Transformer-XL to equip the Transformer with relative position encoding and cached memory for longer context modeling. 
\citet{DBLP:conf/iclr/RaePJHL20} extended the Transformer-XL memory segment to fine-grained compressed memory, which further increases the length of the context and obtains a perplexity of 17.1 on WikiText-103.

Although these works prove that longer context can be helpful for the language modeling task, how to generate better context representations with richer positional information has not been investigated.

On the other hand, large neural LMs trained with a massive amount of text have shown great potential on many NLP tasks, benefiting from the dynamic contextual representations learned from language modeling and other self-supervised pre-training tasks. 
OpenAI GPT~\cite{gpt} and BERT~\cite{DBLP:journals/corr/bert} are two representative models trained with the auto-regressive language modeling task and the masked language modeling task, respectively.
In addition, BERT is also trained with an auxiliary task named next sentence prediction~(NSP).
ALBERT~\cite{DBLP:conf/iclr/albert} then proposed to share parameters across layers of BERT and replaced NSP with sentence order prediction~(SOP). 
According to their experiments, SOP is more challenging than NSP, and MLM together with other downstream tasks can benefit more from replacing NSP with SOP. 
Concurrently to ALBERT, \citet{DBLP:journals/corr/structbert} proposed two auxiliary objectives to provide additional structural information for BERT.

All these powerful pre-trained models encode input tokens with token position encoding, which was first proposed by \citet{DBLP:conf/nips/VaswaniSPUJGKP17} to indicate the position index of the input tokens in the context of machine translation and constituency parsing.
After that, Transformer has been extensively applied in machine translation and other sequence generation tasks~\cite{li-etal-2019-attribute, liu2019text,roller2020recipes}.
However, the input length of language modeling tasks are much longer than these tasks, and simply assigning 0--512 token position embeddings is not enough for LMs to learn the linguistic relationships among these tokens. 
\citet{DBLP:journals/corr/abs-2004-02251} show that incorporating segmentation information with paragraph separating tokens can improve the LM generator~(GPT2) in the context of story generation.
However, compared with punctuation and paragraph breaker, segment position indexes are more straightforward for dot-product self-attention based Transformers. 
In this work, we try to encode segmentation information into the Transformer with the segment-aware position encoding approach.

%% file: sections/conclusion.tex
\section{Conclusion}
In this paper, we propose a novel segment-aware Transformer that can encode richer positional information for language modeling. 
By applying our approach to Transformer-XL, we train a new language model, Segatron-XL, that achieves 17.1 test perplexity on WikiText-103.
Additionally, we pre-trained BERT with our SegaBERT approach and show that our model outperforms BERT on general language understanding, sentence representation learning, and machine reading comprehension tasks.
Furthermore, our SegaBERT-large model outperforms RoBERTa-large on zero-shot STS tasks.
These experimental results demonstrate that our proposed method works on both language models with relative position embeddings and pre-trained language models with absolute position embeddings.

%% file: sections/appendix.tex
\section{Appendix}
\subsection{Self-attention heat maps}
\label{appendix:heatmap}
\begin{figure*}[ht]
  \centering
  \subfigure[SegaBERT-Layer 3]{\includegraphics[width=0.3\textwidth]{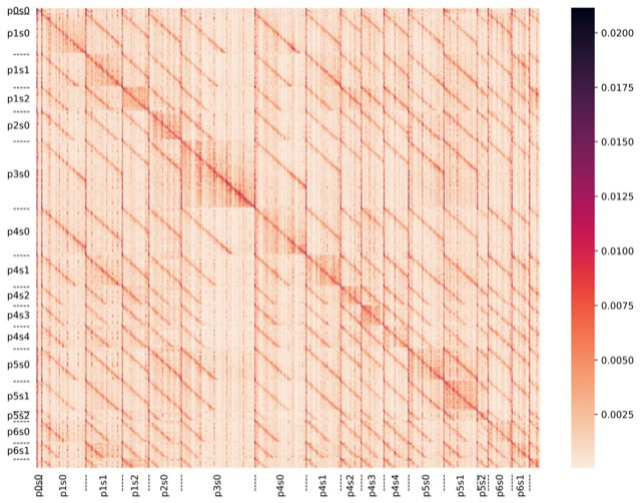}\label{fig: sega 3}}
  \subfigure[BERT-Layer 3]{\includegraphics[width=0.3\textwidth]{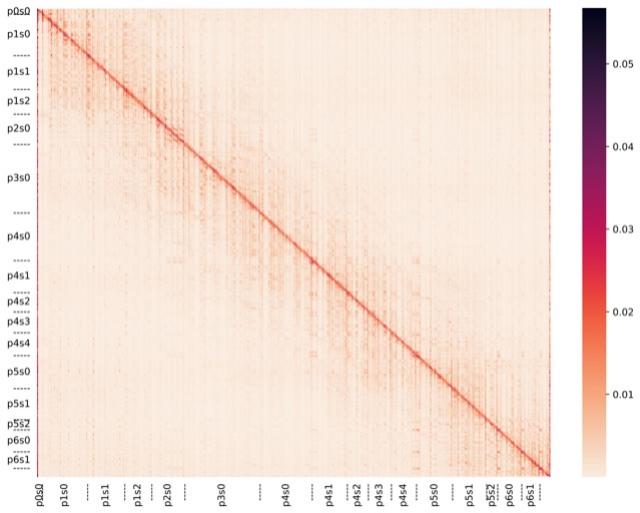}\label{fig: bert 3}}
  
  \caption{Attention heat maps of SegaBERT and BERT base models in the 3rd layer during encoding the first 512 tokens of a Wikipedia article.}
  \label{fig:heatmap3}
\end{figure*}

\begin{figure*}[ht]
  \centering
  \subfigure[SegaBERT-Layer 9]{\includegraphics[width=0.3\textwidth]{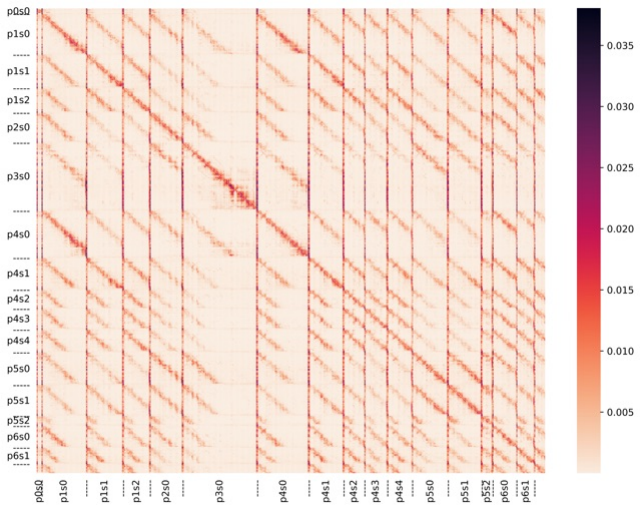}\label{fig: sega 9}}
  \subfigure[BERT-Layer 9]{\includegraphics[width=0.3\textwidth]{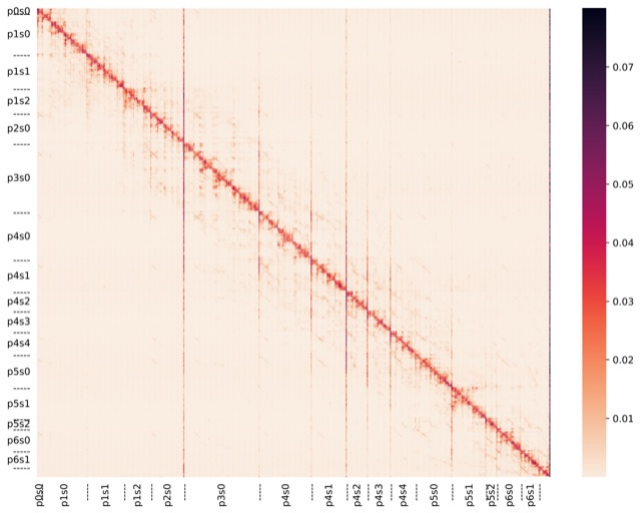}\label{fig: bert 9}}
  
  \caption{Attention heat maps of SegaBERT and BERT base models in the 9th layer during encoding the first 512 tokens of a Wikipedia article.}
  \label{fig:heatmap9}
\end{figure*}

\begin{figure*}[ht]
  \centering
  \subfigure[SegaBERT-Layer 11]{\includegraphics[width=0.3\textwidth]{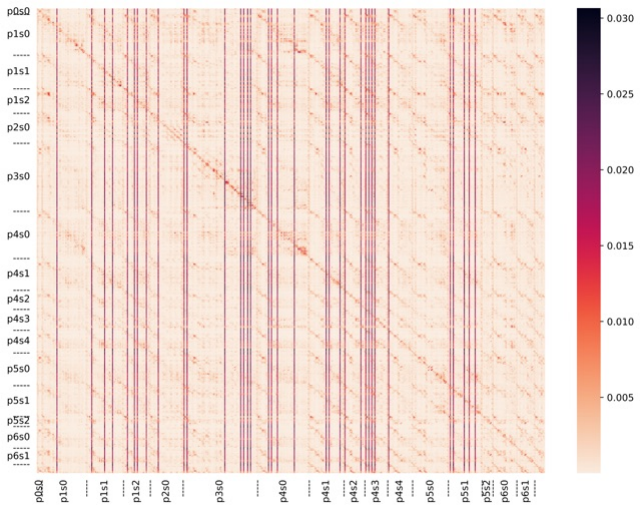}\label{fig: sega 11}}
  \subfigure[BERT-Layer 11]{\includegraphics[width=0.3\textwidth]{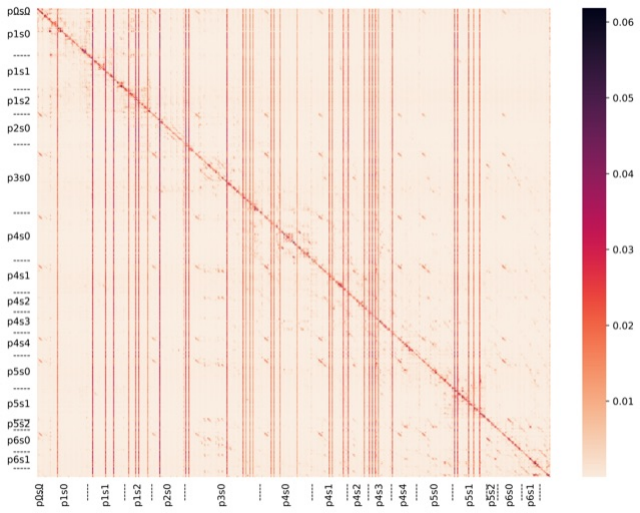}\label{fig: bert 11}}
  
  \caption{Attention heat maps of SegaBERT and BERT base models in the 11th layer during encoding the first 512 tokens of a Wikipedia article.}
  \label{fig:heatmap11}
\end{figure*}

The input article is shown below. 
The actual input is truncated to 512 maximum sequence length after tokenization.
We plot different layer's self attention heat maps in Figure~\ref{fig:heatmap3}, Figure~\ref{fig:heatmap9}, and Figure~\ref{fig:heatmap11}. 

\smallskip \noindent \textbf{Input article:}\\

\fbox{\begin{minipage}{45em}
\fontfamily{Times}\selectfont
    \small
Japanese destroyer Hatsukaze\\

\noindent The \"Kagerō\"-class destroyers were outwardly almost identical to the preceding light cruiser-sized , with improvements made by Japanese naval architects to improve stability and to take advantage of Japan's lead in torpedo technology.
They were designed to accompany the Japanese main striking force and in both day and night attacks against the United States Navy as it advanced across the Pacific Ocean, according to Japanese naval strategic projections.
Despite being one of the most powerful classes of destroyers in the world at the time of their completion, only one survived the Pacific War.\\

\noindent  \"Hatsukaze\", built at the Kawasaki Shipbuilding Corporation, was laid down on 3 December 1937, launched on 24 January 1939 and commissioned on 15 February 1940.\\

\noindent At the time of the attack on Pearl Harbor, \"Hatsukaze\", was assigned to Destroyer Division 16 (Desdiv 16), and a member of Destroyer Squadron 2 (Desron 2) of the IJN 2nd Fleet, and had deployed from Palau, as part of the escort for the aircraft carrier in the invasion of the southern Philippines and minelayer .\\

\noindent In early 1942, \"Hatsukaze\" participated in the invasion of the Netherlands East Indies, escorting the invasion forces for Menado, Kendari and Ambon in January, and the invasion forces for Makassar, Timor and eastern Java in February.
On 27-28 February, \"Hatsukaze\" and Desron 2 participated in the Battle of the Java Sea, taking part in a torpedo attack on the Allied fleet.
During the month of March, Desron 2 was engaged in anti-submarine operations in the Java Sea.
At the end of the month, the squadron escorted the Christmas Island invasion force, then returned to Makassar.
At the end of April, \"Hatsukaze\" sailed to Kure Naval Arsenal for maintenance, docking on 3 May.\\

\noindent On 21 May 1942, \"Hatsukaze\" and Desron 2 steamed from Kure to Saipan, where they rendezvoused with a troop convoy and sailed toward Midway Island.
Due to the defeat of the Carrier Striking Force and loss of four fleet carriers in the Battle of Midway, the invasion was called off and the convoy withdrew without seeing combat.
Desdiv 16 was ordered back to Kure.\\

\noindent On 14 July, \"Hatsukaze\" and Desdiv 16 were reassigned to Desron 10, Third Fleet.
On 16 August, Desron 10 departed Kure, escorting a fleet towards Truk.
On 24 August, Desron 10 escorted Admiral Nagumo's Striking Force in the Battle of the Eastern Solomons.
During September and October, the squadron escorted the fleet patrolling out of Truk north of the Solomon Islands.
On 26 October, in the Battle of the Santa Cruz Islands, the squadron escorted the Striking Force, then escorted the damaged carriers and into Truk on 28 October.
On 4 November, Desron 10 escorted from Truk to Kure, then engaged in training in the Inland Sea, and then escorted \"Zuikaku\" from Truk to the Shortland Islands in January 1943.\\

\noindent On 10 January, while providing cover for a supply-drum transport run to Guadalcanal, \"Hatsukaze\" assisted in sinking the American PT boats \"PT-43\" and \"PT-112.\"
She suffered heavy damage when struck by a torpedo (possibly launched by \"PT-112)\" in the port side; her best speed was 18 knots as she withdrew to Truk, for emergency repairs.
Then she sailed to Kure in April for more extensive repairs.
In September, \"Hatsukaze\" and Desron 10 escorted the battleship from Kure to Truk.
In late September and again in late October, Desron 10 escorted the main fleet from Truk to Eniwetok and back again, in response to American carrier airstrikes in the Central Pacific region.
Between these two missions, \"Hatsukaze\" sortied briefly from Truk in early October 1943 to assist the fleet oiler \"Hazakaya,\" which had been torpedoed by an American submarine.\\

\noindent On 2 November 1943, while attacking an Allied task force off Bougainville in the Battle of Empress Augusta Bay, \"Hatsukaze\" collided with the cruiser .
The collision sheared off her bow, leaving her dead in the water.
\"Hatsukaze\" and the light cruiser were sunk (at position ) by Allied destroyer gunfire.
Of those on board, 164 were killed, including its commanding officer, Lieutenant Commander Buichi Ashida.\\

\noindent \"Hatsukaze\" was removed from the navy list on 5 January 1944."

\end{minipage}}

